\documentclass{article}
\usepackage{spconf,amsmath,graphicx}
\usepackage{cite}
\usepackage{amssymb,amsfonts,amsthm,bm}
\usepackage[usenames,dvipsnames]{xcolor}

\usepackage{booktabs}
\usepackage{graphicx}
\usepackage{multirow}
\usepackage{stfloats}
\usepackage{array} 
\usepackage{caption}
\usepackage{algorithm}
\usepackage{algorithmic}

\definecolor{mygreen}{rgb}{0,0.5,0}
\definecolor{darkblue}{RGB}{0,0,150}
\title{CLIP Multi-modal Hashing: A new baseline}
\name{Jian Zhu$^{1}$, Mingkai Sheng$^{2}$, Mingda Ke$^{2}$, Zhangmin Huang$^{1}$, Jingfei Chang$^{1}$} 
\address{$^{1}$ Zhejiang Lab, China \{qijian.zhu, zmhuang, cjf\_chang\}@zhejianglab.edu.cn\\	$^{2}$  University of Chinese Academy of Sciences, China \{shengmingkai22, kemingda21\}@mails.ucas.ac.cn}
\vspace{-0.35cm}
\begin{document}
\ninept
\maketitle
\ninept
\begin{abstract}
The multi-modal hashing method is widely used in multimedia retrieval. It can fuse multi-source data to generate binary hash code. However, the current multi-modal methods have the problem of low retrieval accuracy. The reason is that the individual backbone networks have limited feature expression capabilities and are not jointly pre-trained on large-scale unsupervised multi-modal data. To solve this problem, we propose a new baseline CLIP Multi-modal Hashing (CLIPMH) method. It uses CLIP model to extract text and image features, and then fuse to generate hash code. CLIP improves the expressiveness of each modal feature. In this way, it can greatly improve the retrieval performance of multi-modal hashing methods. In comparison to state-of-the-art unsupervised and supervised multi-modal hashing methods, experiments reveal that the proposed CLIPMH can significantly enhance performance (Maximum increase of $8.38\%$). CLIP also has great advantages over the text and visual backbone networks commonly used before. The source codes of our CLIPMH is publicly available at: https://github.com/xxx.
\end{abstract}

\section{Introduction}
Multi-modal hashing is one of the important technologies in the field of multimedia retrieval. It is the fusion of multi-modal heterogeneous data to generate hash codes.

The current multi-modal hashing methods have the problem of low retrieval accuracy. The reason is that the backbone networks lack good feature expression capability. For instance, Flexible Multi-modal Hashing (FDH) \cite{zhu:52} and Bit-aware Semantic Transformer Hashing (BSTH) \cite{tan:61} hire a VGG net \cite{simonyan:51} for the image modal. And these methods use a Bag-of-Words model \cite{zhang:53} for the text modal. They are outdated for feature extraction, thus, an update in feature extraction methods is necessary. The fact above results in a degradation of the overall retrieval accuracy for the current multi-modal hashing method.

\begin{figure}
	\centering
	\includegraphics[width=8cm]{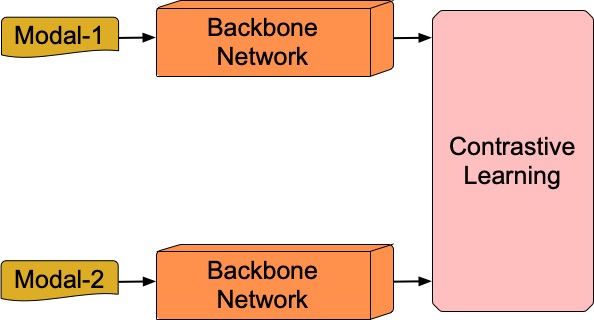}
	\caption{}
	\label{fig:01}
\end{figure}

 In recent years, multi-modal large-scale models have achieved great success. Because these models are trained on large-scale data, they have stronger semantic expression ability. Contrastive Language-Image Pre-training(CLIP) \cite{radford:63} is one of the most representative multi-modal models. However, The application of a multi-modal large model in multi-modal retrieval has not been studied. For the first time, we investigate how CLIP affects the retrieval efficiency of multi-view hashing. As shown in Fig. \ref{fig:01}, it is pre-trained by contrastive learning on large-scale image text data pairs. It has shown exceptional zero-shot or few-shot learning abilities as well as excellent semantic understanding capabilities. The multi-modal field has been greatly changed by CLIP, and more people are beginning to acknowledge that it is superior at multi-modal tasks. Although CLIP has undergone multiple successful trials, a thorough analysis of its effects and performance on multi-modal hashing retrieval has not yet been conducted.

We perform an in-depth study in this work to examine the potential of CLIP on retrieval from multi-modal hashing. We use the CLIP model to extract text and image features. The extracted modal feature data through the CLIP model performs better. It can significantly improve the retrieval performance of multi-modal hashing methods. Compared with the latest state-of-the-art method, the CLIPMH proposed by us has a maximum improvement of $8.38\%$.

Here is a summary of the key contributions of our method:
\begin{itemize}
\item We have studied for the first time the improvement of retrieval performance of multi-modal hashing methods through multi-modal large models.
\item We solve the problem of poor semantic representation in the backbone network of multi-modal hashing methods using the CLIP model.
\item We propose a new multi-modal hashing method termed CLIPMH, which achieves the state-of-the-art result.
\end{itemize}

\begin{figure*}
	\centering
	\includegraphics[width=18cm]{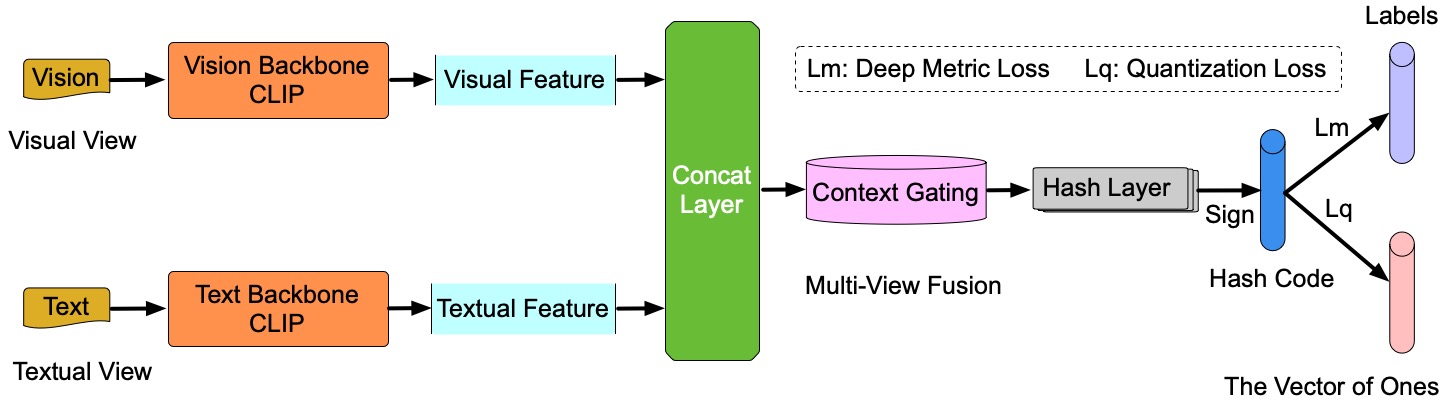}
	\caption{}
	\label{fig:02}
\end{figure*}

\section{The Proposed Methodology}

Deep multi-view hashing network is designed to convert multi-view data into hash code. As shown in Fig. \ref{fig:02}, CLIPMH consists of CLIP backbones, a multi-modal fusion module, and a hash layer. These modules are described in detail below.
\begin{enumerate}
\item \textbf{Vision Backbone:} CLIP \cite{radford:63} is employed to produce visual features. 

\item \textbf{Text Backbone:}  CLIP \cite{radford:63} is utilized to extract text features. 

\item \textbf{Multi-View Fusion Module:}
We employ Context Gating to fuse the concatenated visual and text features. The multi-view fusion module projects the input multi-modal features into a new global representation as:
\begin{equation}
	X_{\text{fusion}} =\sigma(w_{\text{fusion}}X_{\text{concat}}+b_{\text{fusion}})\circ X_{\text{concat}},
\end{equation}
where $X_{\text{concat}} \in \mathbb{R}^{n}$ is the multi-view feature vector, $\sigma$ is the element-wise sigmoid activation, and $\circ$  is the element-wise multiplication.  $w_{\text{fusion}} \in \mathbb{R}^{n \times n}$  and $b_{\text{fusion}} \in \mathbb{R}^{n}$ are trainable parameters. The vector of weights $\sigma(w_{\text{fusion}}X_{\text{concat}}+b_{\text{fusion}}) \in [0, 1]$ represents a set of learned gates applied to the individual dimensions of the input feature $X_{\text{concat}}$.

\item \textbf{Hash Layer:} A linear layer with a $\tanh$ activation is hired as the hash layer, which can be represented as $h_{\text{k-bit}} = \text{sgn}[\tanh(w_{\text{hash}}X_{\text{fusion}}+b_{\text{hash}})]$,
where $sgn$ represents the signum function. $w_{\text{hash}} \in \mathbb{R}^{n \times n}$  and $b_{\text{hash}} \in \mathbb{R}^{n}$ are trainable parameters. The output has the same number of dimensions as the hash code. 
\end{enumerate}

\begin{table*}
	\centering
        \caption{General statistics of three datasets. The dataset size, number of categories, and feature dimensions are included.}
	\begin{tabular}{llllllll}
		\toprule[1pt]
		Dataset   & Training Size & Retrieval Size & Query Size & Categories&Visual Embedding & Textual Embedding \\ \midrule[0.8pt]
		MIR-Flickr25K & 5000  & 17772 & 2243    & 24&512-D & 512-D \\
  NUS-WIDE & 21000  & 193749 & 2085    & 21&512-D &512-D\\
		MS COCO & 18000  & 82783 & 5981    & 80&512-D &512-D\\
		
		\bottomrule[1pt]
	\end{tabular}
	
	\label{Tab:01}
\end{table*}

\begin{table*}
	\setlength{\tabcolsep}{2pt}
	\centering
	\caption{The comparable mAP results on MIR-Flickr25K, NUS-WIDE, and MS COCO. The best results are bolded, and the second-best results are underlined. The * indicates that the results of our method on this dataset are statistical significance.}
	\resizebox{\textwidth}{!}{\begin{tabular}{llllllllllllll}
		\toprule[1pt]
		\multicolumn{1}{c}{\multirow{2}{*}{Method}} & \multicolumn{1}{c}{\multirow{2}{*}{Ref.}} & \multicolumn{4}{c}{MIR-Flickr25K*}    & \multicolumn{4}{c}{NUS-WIDE*}      & \multicolumn{4}{c}{MS   COCO*}       \\  \cmidrule(r){3-6}  \cmidrule(r){7-10}  \cmidrule(r){11-14}
		\multicolumn{1}{c}{}                         & \multicolumn{1}{c}{}                      & 16 bits & 32 bits & 64 bits & 128 bits & 16 bits & 32 bits & 64 bits & 128 bits & 16 bits & 32 bits & 64 bits & 128 bits \\ \midrule[0.8pt]
		MFH                                          & TMM13                                     & 0.5795 & 0.5824 & 0.5831 & 0.5836  & 0.3603 & 0.3611 & 0.3625 & 0.3629  & 0.3948 & 0.3699 & 0.3960  & 0.3980   \\
		MAH                                          & TIP15                                     & 0.6488 & 0.6649 & 0.6990  & 0.7114  & 0.4633 & 0.4945 & 0.5381 & 0.5476  & 0.3967 & 0.3943 & 0.3966 & 0.3988  \\
		MVLH                                         & MM15                                      & 0.6541 & 0.6421 & 0.6044 & 0.5982  & 0.4182 & 0.4092 & 0.3789 & 0.3897  & 0.3993 & 0.4012 & 0.4065 & 0.4099  \\
		MvDH                                         & TIST18                                    & 0.6828 & 0.7210  & 0.7344 & 0.7527  & 0.4947 & 0.5661 & 0.5789 & 0.6122  & 0.3978 & 0.3966 & 0.3977 & 0.3998  \\ \midrule[0.8pt]
		MFKH                                         & MM12                                      & 0.6369 & 0.6128 & 0.5985 & 0.5807  & 0.4768 & 0.4359 & 0.4342 & 0.3956  & 0.4216 & 0.4211 & 0.4230  & 0.4229  \\
		DMVH                                         & ICMR17                                    & 0.7231 & 0.7326 & 0.7495 & 0.7641  & 0.5676 & 0.5883 & 0.6902 & 0.6279  & 0.4123 & 0.4288 & 0.4355 & 0.4563  \\
		FOMH                                         & MM19                                      & 0.7557 & 0.7632 & 0.7564 & 0.7705  & 0.6329 & 0.6456 & 0.6678 & 0.6791  & 0.5008 & 0.5148 & 0.5172 & 0.5294  \\
		FDMH                                         & NPL20                                     & 0.7802 & 0.7963 & 0.8094 & 0.8181  & 0.6575 & 0.6665 & 0.6712 & 0.6823  & 0.5404 & 0.5485 & 0.5600   & 0.5674  \\
		DCMVH                                        & TIP20                                     & 0.8097 & 0.8279 & 0.8354 & 0.8467  & 0.6509 & 0.6625 & 0.6905 & 0.7023  & 0.5387 & 0.5427 & 0.5490  & 0.5576  \\
		SAPMH                                        & TMM21                                      & 0.7657 & 0.8098 & 0.8188 & 0.8191  & 0.6503 & 0.6703 & 0.6898 & 0.6901  & 0.5467 & 0.5502 & 0.5563 & 0.5672  \\
		FGCMH                                        & MM21                                      & 0.8173 & 0.8358 & 0.8377 & 0.8606  & 0.6677 & 0.6874 & 0.6936 & 0.7011  & 0.5641 & 0.5273 & 0.5797 & 0.5862 \\
  BSTH & SIGIR22  &0.8145&0.8340&0.8482&0.8571                                     & 0.6990 & 0.7340 & 0.7505 & 0.7704  & 0.5831 & 0.6245 & 0.6459 & 0.6654  \\
  DMMVH                                         & ICME23                                         &\underline{0.8587} & \underline{0.8707} & \underline{0.8798} & \underline{0.8827} & \underline{0.7714} & \underline{0.7820} & \underline{0.7879} & \underline{0.7916}  &\underline{0.6716} & \underline{0.7030} & \underline{0.7122} & \underline{0.7244}\\\midrule[0.8pt]
		CLIPMH                                         & Proposed                                         &\textbf{0.8862} & \textbf{0.8921} & \textbf{0.8956} & \textbf{0.8975} & \textbf{0.7802} & \textbf{0.7986} & \textbf{0.8029} & \textbf{0.8085}  &\textbf{0.6806} & \textbf{0.7450} & \textbf{0.7693} & \textbf{0.8082} \\
		\bottomrule[1pt]
	\end{tabular}}

	\label{Tab:02}
\end{table*}

\section{Experiments}
\subsection{Evaluation Datasets and Metrics}
In the experiments, we evaluate the performance of the proposed CLIPMH model on large-scale multimedia retrieval tasks. We utilize three well-known datasets: MIR-Flickr25K \cite{huiskes:20}, NUS-WIDE \cite{chua:22}, and MS COCO \cite{lin:21}. These datasets have gained widespread usage for evaluating the performance of multimedia retrieval systems. The mean Average Precision (mAP) is employed as the evaluation metric. Table \ref{Tab:01} provides a summary of the dataset statistics used in the experiments.

\subsection{Baseline}
To evaluate the retrieval metric, we compare the proposed CLIPMH method with thirteen multi-view hashing methods, including four unsupervised methods (e.g., Multiple Feature Hashing (MFH) \cite{song:9}, Multi-view Alignment Hashing (MAH) \cite{liu:10}, Multi-view Latent Hashing (MVLH) \cite{shen:6}, and Multi-view Discrete Hashing (MvDH) \cite{shen:11}) and nine supervised methods (e.g., Multiple Feature Kernel Hashing (MFKH) \cite{liu:7}, Discrete Multi-view Hashing (DMVH) \cite{yang:12}, Flexible Discrete Multi-view Hashing (FDMH) \cite{liu:23}, Flexible Online Multi-modal Hashing (FOMH) \cite{lu:24}, Deep Collaborative Multi-View Hashing (DCMVH) \cite{zhu:17}, Supervised Adaptive Partial Multi-view Hashing (SAPMH) \cite{zheng:25}, Flexible Graph Convolutional Multi-modal Hashing (FGCMH) \cite{lu:18}, Bit-aware Semantic Transformer Hashing (BSTH) \cite{tan:61} and Deep Metric Multi-View Hashing (DMMVH) \cite{zhu:62}).

\subsection{Analysis of Experimental Results}

The mAP results are presented in Table \ref{Tab:02}. The results demonstrate that the proposed CLIPMH method outperforms all the compared multi-view hashing methods by a significant margin. Specifically, when compared to the current state-of-the-art multi-modal hashing method  DMMVH \cite{zhu:62}, our method achieves an average mAP improvement of $2.00\%$, $1.43\%$, and $4.80\%$ on the MIR-Flickr25K, NUS-WIDE, and MS COCO datasets, respectively. These superior results can be attributed to three main factors:

\begin{itemize}
\item We use the CLIP model to build a multi-modal hashing method. 
\item The CLIP model extracts image features and enhances semantic expression. 
\item The CLIP model extracts text features and enhances semantic expression. 
\end{itemize}

\section{Conclusion and Future Work}
In this paper, we propose a new multi-modal hashing method termed CLIPMH. Experiments show that our method achieves state-of-the-art results. CLIP model is used to solve the problem that backbone network extraction feature semantic expression is insufficient in the multi-modal hashing method. We provide a new baseline method for the multimodal hashing domain. In the future, we will study more application issues of multi-modal large models in the field of multimedia retrieval.

\vspace{-9pt}
  
\ninept 
\bibliographystyle{IEEEtran}
\bibliography{IEEEabrv,myabrv_new,my_reference}
\end{document}